\documentclass[conference,9pt]{IEEEtran}
\IEEEoverridecommandlockouts
\usepackage{cite}
\usepackage{amsmath,amssymb,amsfonts}
\usepackage{algorithmic}
\usepackage{graphicx}
\usepackage{textcomp}
\usepackage{xcolor}
\def\BibTeX{{\rm B\kern-.05em{\sc i\kern-.025em b}\kern-.08em
    T\kern-.1667em\lower.7ex\hbox{E}\kern-.125emX}}

\usepackage{algorithm}
\usepackage{algorithmic}

\makeatletter
\renewcommand{\ALG@name}{\textbf{Algorithm}}
\makeatother

\usepackage{amsmath}
\usepackage{color}
\usepackage{multirow}
\usepackage{amssymb}
\usepackage{balance}

\newtheorem{definition}{Definition}

\definecolor{Xiang}{rgb}{1,0,0}

\newcommand{\nop}[1]{}

\begin{document}

\title{
FitLight: Federated Imitation Learning for Plug-and-Play Autonomous Traffic Signal Control
}

\author{\IEEEauthorblockN{Yutong Ye}
\IEEEauthorblockA{\textit{East China Normal University} \\
Shanghai, China \\
ytye@stu.ecnu.edu.cn}
\and
\IEEEauthorblockN{Yingbo Zhou}
\IEEEauthorblockA{\textit{East China Normal University} \\
Shanghai, China \\
52215902009@stu.ecnu.edu.cn}
\and
\IEEEauthorblockN{Xiao Du}
\IEEEauthorblockA{\textit{East China Normal University} \\
Shanghai, China \\
52265902007@stu.ecnu.edu.cn}
\and
\IEEEauthorblockN{Zhusen Liu}
\IEEEauthorblockA{\textit{East China Normal University} \\
Shanghai, China \\
52184501023@stu.ecnu.edu.cn}
\and
\IEEEauthorblockN{Hao Zhou}
\IEEEauthorblockA{\textit{East China Normal University} \\
Shanghai, China \\
hzhou6@kent.edu}
\and
\IEEEauthorblockN{Xiang Lian}
\IEEEauthorblockA{\textit{Kent State University} \\
Kent, United States \\
xlian@kent.edu}
\and
\IEEEauthorblockN{Mingsong Chen}
\IEEEauthorblockA{\textit{East China Normal University} \\
Shanghai, China \\
mschen@sei.edu.cn}
}

\maketitle

\begin{abstract}
Although Reinforcement Learning (RL)-based Traffic Signal Control (TSC) methods have been extensively studied, their practical applications still raise some serious issues such as high learning cost and poor generalizability. This is because the ``trial-and-error'' training style makes RL agents extremely dependent on the specific traffic environment, which also requires a long convergence time. To address these issues, we propose a novel Federated Imitation Learning (FIL)-based framework for multi-intersection TSC, named FitLight, which allows RL agents to plug-and-play for any traffic environment without additional pre-training cost. Unlike existing imitation learning approaches that rely on pre-training RL agents with demonstrations, FitLight allows real-time imitation learning and seamless transition to reinforcement learning. Due to our proposed knowledge-sharing mechanism and novel hybrid pressure-based agent design, RL agents can quickly find a best control policy with only a few episodes. Moreover, for resource-constrained TSC scenarios, FitLight supports model pruning and heterogeneous model aggregation, such that RL agents can work on a micro-controller with merely 16{\it KB} RAM and 32{\it KB} ROM. Extensive experiments demonstrate that, compared to state-of-the-art methods, FitLight not only provides a superior starting point but also converges to a better final solution on both real-world and synthetic datasets, even under extreme resource limitations.
\end{abstract}

\begin{IEEEkeywords}
Autonomous System, Traffic Signal Control, Federated Imitation Learning, Hybrid Pressure
\end{IEEEkeywords}

\vspace{-1ex}
\section{Introduction}
\vspace{-1ex}
With the rapid development of cities and the rapid growth of population, more and more cities are suffering from heavy traffic congestion, resulting in various serious problems, such as economic losses, increasing commuting costs, and environmental pollution. The \textit{Traffic Signal Control} (TSC) has attracted widespread attention as a promising solution to traffic congestion \cite{chen2013neuromorphic,waszecki2017automotive,chang2020cps}. In most real-world applications, the control policies are rule-based (e.g., FixedTime \cite{koonce2008traffic}, GreenWave \cite{torok1996green}, SCOOT \cite{hunt1982scoot}, and SCATS \cite{pr1992scats}) that follow some pre-defined rules of traffic plan. To better deal with dynamic traffic scenarios, some adaptive methods have been proposed (e.g., MaxPressure \cite{varaiya2013max}, MaxQueue \cite{zhang2021knowledge}, and SOTL \cite{cools2013self}), which control traffic in a heuristic manner. Due to the prosperity of Artificial Intelligence (AI) and Internet of Things (IoT) technologies, using Reinforcement Learning (RL), to control traffic signals has been a promising way.

Although RL-based methods can achieve better control performance, their usage is greatly restricted by the issues of high learning cost and poor generalizability. This is because of the "trial-and-error" learning style of RL agents, which requires the RL agent to make a large number of attempts in a specific traffic environment to gradually learn the control strategy. Worse still, as the size of the road network and the number of RL agents increase, the size of the policy space will also increase exponentially, which will make it extremely difficult to find the optimal control strategy in multi-intersection scenarios and require a lot of training costs. Therefore, how to effectively improve both the training efficiency and generalization ability is becoming a major challenge in the application of RL-based TSC methods.

\begin{figure}[h]  
    \centerline{\includegraphics[width=0.9\columnwidth]{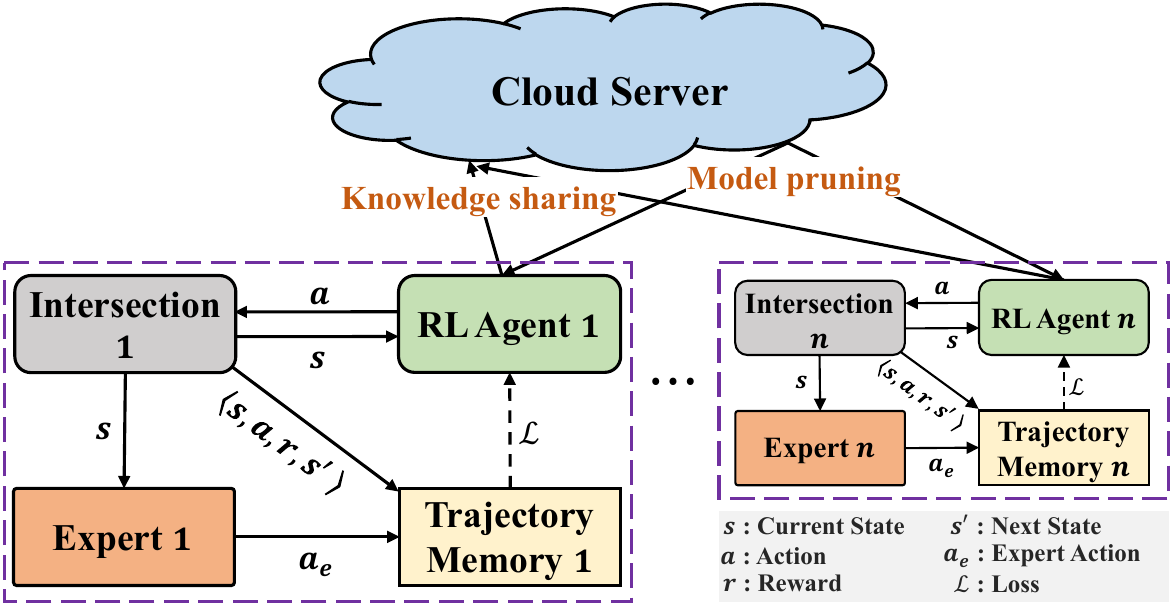}}
    \vspace{-2ex}
    \caption{Framework of FitLight.}
    \vspace{-4ex}
    \label{fig:overall}
\end{figure}

To tackle this problem, in this paper, we propose a novel Federated Imitation Learning (FIL)-based framework named FitLight for efficient and effective multi-intersection TSC, which enables RL agents plug-and-play for different traffic environments without additional pre-training cost. In other words, after obtaining a very high-quality solution in the first episode, FitLight can quickly converge to a better final control strategy. As shown in Figure~\ref{fig:overall}, FitLight is built on a cloud-edge framework consisting of one cloud server and multiple edge nodes (i.e., RL agent and its corresponding intersection). Unlike existing methods that either train RL agents directly within the traffic environment or employ imitation learning over pre-collected data for pre-training, FitLight seamlessly integrates imitation learning into the reinforcement learning process. This integration allows the RL agent to achieve a high-quality initial solution in the first episode due to the supervision of imitation learning. Subsequently, due to our novel hybrid pressure-based agent design, the RL agent seamlessly transitions into the reinforcement learning phase, ultimately converging to an even better control strategy. Moreover, FitLight's support for model pruning and heterogeneous model aggregation ensures that RL agents can be deployed in resource-constrained TSC scenarios. In summary, this paper makes the following four major contributions:

\vspace{-0.5ex}
\begin{itemize}
    \item We propose a novel Federated Imitation Learning (FIL)-based framework that can plug and play for any traffic environment without additional pre-training costs.
    \item We introduce an imitation learning mechanism combined with a hybrid pressure-based agent design, enabling real-time imitation learning and smooth transitions to reinforcement learning, allowing RL agents to quickly achieve a high-quality solution in the first episode.
    \item We propose a federated learning-based knowledge sharing mechanism with the support of heterogeneous model aggregation, which improves learning efficiency and enables our FitLight to work in TSC scenarios with extremely limited resources.
    \item Extensive experiments on various synthetic and real-world datasets show the superiority of our FitLight in terms of average travel time and convergence rate.
\end{itemize}

\begin{figure*}[t]
    \centerline{\includegraphics[width=1.7\columnwidth]{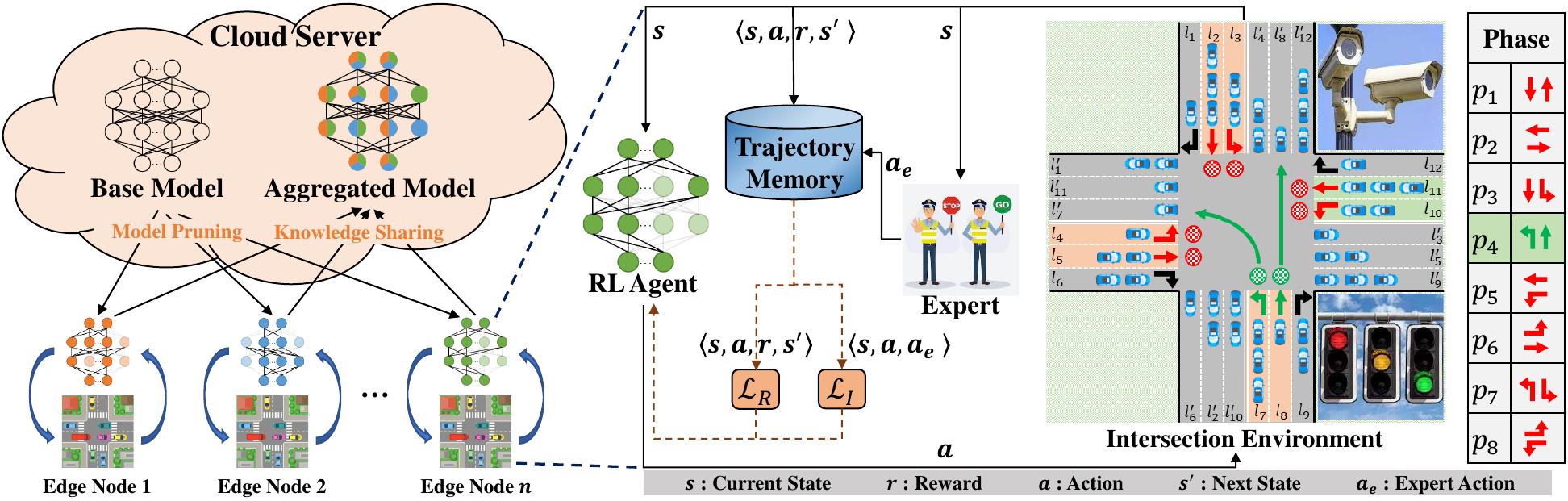}}
    \vspace{-2ex}
    \caption{Architecture and workflow of FitLight.}
    \vspace{-4ex}
    \label{fig:detail}
\end{figure*}

\vspace{-1ex}
\section{Related Work}
\vspace{-1ex}
To improve the performance of TSC, various methods based on RL have been proposed. For example, PressLight \cite{wei2019presslight}, CoLight \cite{wei2019colight}, MPLight \cite{chen2020toward}, MetaLight \cite{zang2020metalight}, and RTLight \cite{ye2023brief} performed TSC optimization based on the concept of pressure from the Max Pressure (MP) control theory \cite{varaiya2013max} to design the state and reward. Unlike these methods, IPDALight \cite{zhao2022ipdalight} proposed a new concept named intensity, which investigates both the speed of vehicles and the influence of neighboring intersections. To reflect the fairness of individual vehicles, FairLight \cite{ye2022fairlight} and FELight~\cite{du2024felight} considered the relationship between waiting time and driving time, and the extra waiting time of vehicles, deceptively. To exploit the cooperation among RL agents in the road network, FedLight \cite{ye2021fedlight} and RTLight \cite{ye2023brief} adopt federated reinforcement learning to share knowledge. HiLight \cite{xu2021hierarchically} cooperatively controls traffic signals to directly optimize average travel time by using hierarchical reinforcement learning. UniLight \cite{jiang2022multi} uses a universal communication form between intersections to implement cooperation. However, the RL agents of these methods are trained from the randomly initialized models, resulting in a long training time before obtaining the final control strategy.

To improve learning efficiency, imitation learning \cite{agarwal2019reinforcement,argall2009survey,hussein2017imitation,osa2018algorithmic} that makes the RL agent learn from the expert demonstration is a promising way. Currently, imitation learning can be divided into two categories: behavioral cloning \cite{bain1995framework,pomerleau1991efficient} and adversarial imitation learning \cite{abbeel2004apprenticeship,syed2007game,ziebart2008maximum}, both of which have been applied in TSC. Specifically, DemoLight \cite{xiong2019learning} is a behavioral cloning-based method that reduces the imitation learning task to a common classification task \cite{ross2010efficient,syed2010reduction} by minimizing the action difference between the agent strategy and the expert strategy. However, since this method is trained in the single-intersection environment and relies on the pre-collected expert trajectory from the same environment, it cannot be applied to multi-intersection scenarios and is very specific to the training environment. On the other hand, as an adversarial imitation learning-based method, InitLight \cite{ye2023initlight} uses a generative adversarial framework to learn expert's behaviors, where the discriminator iteratively differentiates between pre-collected expert and agent trajectories (generated through real-time agent-environment interactions). Although InitLight can use trajectories from different environments to train RL agents, it still needs a pre-training process to obtain an initial model.

To the best of our knowledge, FitLight is the first federated imitation learning framework for TSC to enable RL agents to plug-and-play for any traffic environment without additional pre-training cost, where the RL agent can achieve a high-quality initial solution in the first episode and then converge to an even better final result.

\vspace{-1ex}
\section{Our FitLight Approach}
\label{our_method}
\vspace{-1ex}
To make RL agents plug-and-play in different traffic scenarios without pre-training, we design a novel FitLight approach, based on a cloud-edge architecture, where the cloud server is used for knowledge sharing among intersections, and each intersection equips an RL agent and an expert strategy. Figure~\ref{fig:detail} details the FitLight components and workflow. In our approach, the federated imitation learning framework consists of a cloud server and several edge nodes. The cloud server first dispatches models pruned from a base model to edge nodes, and then shares knowledge among different intersections by aggregating heterogeneous models during the RL training. For each edge node, we deploy an RL agent to monitor traffic dynamics using connected sensors, e.g., cameras, make the traffic signal control decision, and update network parameters. Once capturing the current traffic state $s$, the RL agent will choose one best action $a$ to control traffic lights. Meanwhile, the expert strategy also gives a decision $a_e$, which will be stored as the label of the current state for imitation learning. This expert guidance enables the RL agent to quickly identify a high-quality solution. We will give the details of our approach in the following subsections.

\vspace{-1ex}
\subsection{Intersection Modeling}
\vspace{-0.5ex}
The right part of Figure~\ref{fig:detail} shows an intersection example with three components, i.e., arrival and departure lanes, directed roads, and control phase setting:
\vspace{-0.5ex}
\begin{itemize}
    \item {\bf Arrival and Departure Lanes:} The intersection consists of a set of arrival lanes $L_a=\{l_1,l_2,\cdots,l_{12}\}$ and a set of departure lanes $L_d=\{l'_1,l'_2,\cdots,l'_{12}\}$, where vehicles can enter and exit the intersection, respectively.
    \item {\bf Directed Roads:} Based on direction marks at the end of each arrival lane, we define a directed road as $(l_a,l_d),l_a\in L_a,l_d\in L_d$, where $l_d$ is the departure lane indicated by the direction mark on the ground of $l_a$. For example, $(l_7,l'_7)$ and $(l_8,l'_8)$ are two directed roads.
    \item {\bf Control Phases:} According to common sense, the vehicles turning right are not restricted by traffic. Therefore, we design a set of eight feasible control phases $P=\{p_1, p_2, \cdots, p_8\}$, which are obtained by combining 2 of 8 directed roads and indicate the rights-of-way signaled to vehicles by traffic lights. For example, the intersection on the right side of Figure~\ref{fig:detail} shows a scenario with control phase $p_4$ enabled, where the vehicles on lane $l_7$ can turn left to enter lane $l'_7$ and the vehicles on lane $l_8$ can go straight to enter lane $l'_8$. 
    Note that, the number of control phases is fixed, regardless of the number of lanes.
\end{itemize}
\vspace{-0.5ex}

\vspace{-1ex}
\subsection{Hybrid Pressure}
\vspace{-0.5ex}
Unlike most existing works that utilize pressure from MP control theory to model traffic dynamics, in this paper, we introduce a novel concept of Hybrid Pressure (HP) for the design of RL agents.
Specifically, HP considers more dynamics from the individual vehicle level to intersection level rather than only the number of vehicles. Therefore, it can be used for more accurate modeling of RL elements.

\begin{definition}
\label{def:hp_vehicle}
    (Hybrid Pressure of a Vehicle). The hybrid pressure of a vehicle $veh$ on a lane of an intersection $hp_{veh}$ is defined as: 
    \vspace{-1ex}
    \begin{equation}
        hp_{veh}=log(1+\frac{l_{max}-d}{l_{max}}+\frac{v_{max}-v}{v_{max}}+\frac{wt_{veh}}{dt_{veh}}),
        \label{eq:hp_vehicle}
        \vspace{-1ex}
    \end{equation}
    where $l_{max}$ is the lane length, $d$ indicates the distance between the vehicle and the intersection, $v_{max}$ is the maximum speed of the lane, $v$ is the current speed of the vehicle, $wt_{veh}$ and $dt_{veh}$ are the overall waiting time and driving time of the vehicle along its route so far from the time when it entering the traffic network, respectively. 
    Note that we normalize the distance and speed by $l_{max}$ and $v_{max}$ to constrain their ranges. 
    Moreover, we use $log(\cdot)$ to smooth the absolute value of $h_{veh}$ and plus 1 to make sure $h_{veh}$ is always greater than 0.
\end{definition}

We use the HP of a vehicle to indicate the traffic priority when the vehicle arrives at an intersection. According to Definition~\ref{def:hp_vehicle}, we can find that the vehicles with a shorter distance to the intersection, a slower speed, and a longer cumulative waiting time will have higher $h_{veh}$ value, i.e., have greater priority for the right of way. When waiting at some intersection, the waiting time of a vehicle increases cumulatively, which results in an increase in the corresponding lane HP value. Along with the increasing lane HP values, the vehicles on some feeder roads can move eventually. Due to the elegant combination of individual vehicles' features, HP can more accurately reflect the traffic dynamics.

\begin{definition}
\label{def:hp_road}
    (Hybrid Pressure of a Directed Road). The hybrid pressure of a directed road $(l_a,l_d)$  is defined as the difference in hybrid pressure between the arrival and departure lanes, where a lane's hybrid pressure is the sum of all vehicles' HP on that lane.
    \vspace{-1ex}
    \begin{equation}
        hp_{(l_a,l_d)}=\sum_{veh\in l_a}hp_{veh}-\sum_{veh\in l_d}hp_{veh}.
        \label{eq:hp_road}
    \end{equation}
\end{definition}
\vspace{-1ex}
To denote the hybrid pressure of all the vehicles on a lane, Definition~\ref{def:hp_road} defines the HP of a directed road $(l_a,l_d)$. Since the hybrid pressure of a lane is in a summation form, it implicitly reflects the number of vehicles on on both $l_a$ and $l_d$. Therefore, $hp_{(l_a, l_d)}$ not only reflects the status of individual vehicles but also captures the imbalance in traffic conditions between the upstream and downstream lanes. In our approach, the TSC controller tends to allow the directed road with higher HP values to move first. 

\begin{definition}
\label{def:hp_intersection}
    (Hybrid Pressure of an Intersection). The hybrid pressure of an intersection $I$ equals the difference between the arrival and departure lanes' HP values, i.e., 
    \vspace{-1ex}
    \begin{equation}
        hp_{I}=\sum_{l\in L_a}hp_{l}-\sum_{l\in L_d}hp_{l},
        \label{eq:hp_intersection}
    \end{equation}
\end{definition}
\vspace{-1ex}

Definition~\ref{def:hp_intersection} presents how to calculate the FI for an intersection, which can be used to evaluate the overall traffic pressure faced by the intersection. From this definition, we can find that the hybrid pressure of the intersection can approximately reflect the imbalance of upstream and downstream traffic status at the intersection. Therefore, similar to the MP control theory, if the HP of the intersection can be controlled at a low level, the throughput of vehicles crossing this intersection will be maximized.

\subsection{Edge Node Design}
In our approach, the traffic lights of each edge node are controlled by a Proximal Policy Optimization (PPO) \cite{schulman2017proximal} agent. Unlike most existing works that directly train the agent by reinforcement learning, we deploy an expert algorithm to guide the agent for efficient convergence by imitation learning, which can make the RL agent find a high-quality solution in the first episode. 

\noindent{\bf Expert Algorithm.} Based on the concept of hybrid pressure, we design a simple but effective control heuristic named MaxHP, which greedily selects the control phase with the maximum HP values. Similar to the classic MP-based heuristic control method MaxPressure \cite{varaiya2013max}, by allowing vehicles in the lane with the largest HP value to pass, MaxHP can reduce the HP value of the intersection.

\noindent{\bf Agent Design.} In this paper, we design the key elements of the PPO agent by using the proposed HP concept, i.e.,

\begin{itemize}
    \item {\bf State:} State is the information of the intersection captured by the agent as its own observation for phase selection. 
    Take the standard intersection in Figure~\ref{fig:detail} as an example,
    the state includes the HP of all directed lanes (i.e., $hp_{l_1,l'_1},hp_{l_2,l'_2},\cdots,hp_{l_{12},l'_{12}}$ and the current control phase (i.e., $p_4$).

    \item {\bf Action:} Based on the observed current traffic state, the PPO agent needs to choose one best control phase to maximize the throughput of the intersection. For the intersection example in Figure~\ref{fig:detail}, the PPO agent has 8 permissible control phases (i.e., $p_1,\cdots,p_8$).

    \item {\bf Reward:} Once an action is completed, the environment will return a reward to the agent. The reward mechanism plays an important role in the RL learning process. It is required that a higher reward needs to imply a better action choice. As mentioned in Definition~\ref{def:hp_intersection}, to encourage the agent to maximize the throughput of the intersection by minimizing the hybrid pressure of the intersection $hp_{I}$, in this paper, we define the reward as $r=-hp_I$.
\end{itemize}

A PPO agent consists of two trainable networks, i.e., Actor $\theta_A$ and Critic $\theta_C$, where $\theta$ is the model parameter. The Actor model is responsible for learning the policy, and determining which action to take given the current state. The Critic model, on the other hand, serves as a value estimator, assessing whether the action selected by the Actor will lead to an improved state in the traffic environment. Therefore, the feedback from the Critic model can also be used to optimize the Actor model. In this paper, since we also utilize imitation learning to guide the agent training, as shown in the right part of Figure~\ref{fig:detail}, there are two loss functions from reinforcement learning $L_R$ and imitation learning $L_I$. To calculate these loss values for optimizations, a mini-batch of trajectory samples is collected from the agent trajectory memory, where each sample is a quintuple $\langle s,a,a_e,r,s' \rangle$.

First, the reinforcement learning loss $L_R$ includes the losses of both the Critic model $L_C$ and the Actor model $L_A$. 
Note that, the reinforcement learning of PPO requires that trajectory samples in a mini-batch be continuous.

\underline{\it Critic Model.} 
We optimize the Critic model by:
\vspace{-1ex}
\begin{equation}
    L_C=\mathbb{E}[|\theta_C(s_{t})_{target}-\theta_C(s_{t})|],
    \label{eq:closs}
    \vspace{-1ex}
\end{equation}

where $\mathbb{E}$ is an operator to calculate the empirical average over a mini-batch of samples, and $\theta_C(s_{t})_{target}$ can be calculated as $\theta_C(s_{t})_{target}=r_{t+1}+\gamma \cdot C(s_{t+1})$ by using the Temporal-Difference (TD) algorithm \cite{tesauro1995temporal} to estimate the target value.

\underline{\it Actor Model.} 
As a policy gradient-based RL algorithm, the objective of the Actor model is formulated as follows:
\vspace{-1ex}
\begin{equation}
    L_A=\mathbb{E}[min(R_t,clip(R_t,1-\sigma,1+\sigma))A_t],
    \label{eq:aloss}
    \vspace{-1ex}
\end{equation}
where $R_t=\frac{\theta_A(a_t|s_t)}{\theta^{old}_{A}(a_t|s_t)}$ is the importance sampling that obtains the expectation of samples under the new Actor model $\theta_A$ we need to update, $A_t$ is an estimated value of the advantage function at time step $t$, and $\sigma$ is the clipping parameter that restricts the upper/lower bounds in the $clip(\cdot)$ function to stabilize the updating process. Note that, the samples are gathered from an old Actor model $\theta^{old}_A$. 
The advantage function $A_t$ is computed with the Generalized Advantage Estimator (GAE) \cite{schulman2015high} as follows:
\vspace{-1ex}
\begin{equation}
\begin{aligned}
    A_{t}=\delta_t+(\gamma \lambda)\delta_{t+1}+(\gamma \lambda)^{2}\delta_{t+2}\\+\cdots+(\gamma \lambda)^{|B|-t+1}\delta_{|B|-1},
\end{aligned}
\vspace{-1ex}
\end{equation}
where $\gamma \in [0,1]$ is the discount factor of future rewards, $\lambda \in [0,1]$ is the GAE parameter, $|B|$ is the batch size of the sampled mini-batch, and $\delta_t=r_t+\gamma \theta_C(s_{t+1})-
\theta_C(s_t)$.

Moreover, the Actor model also has a loss from imitation learning, which is used to guide the agent's behavior. 

\underline{\it Imitation Learning.} 
In our approach, we use MaxHP as the expert algorithm to label the state $s$ by selecting the corresponding action $a_e$. This labeling process enables us to apply supervised learning to minimize the discrepancy between the Actor’s actions and the expert’s actions. Since the control phases are discrete actions, we employ the cross-entropy loss function \cite{shannon1948mathematical,zhang2018generalized} to handle this multi-class classification task:
\vspace{-2ex}
\begin{equation}
    L_I=-\sum_{i=1}^{|P|}a_{e_i} log(a_i),
    \label{eq:iloss}
    \vspace{-1ex}
\end{equation}
where $|P|$ is the number of classes (control phases), $a_{e_i}$ is the indicator variable of the label (i.e., the action chosen by the expert algorithm) that is encoded by a one-hot vector, $a_i$ is the prediction probability of the $i$-th action given by the Actor model.

Finally, considering the balance of exploitation and exploration for the RL agent training, we use a balance factor $\alpha$ to adjust the weighting of different losses:
\vspace{-1.5ex}
\begin{equation}
    L=\alpha(L_C+L_A)+(1-\alpha)L_I,
    \label{eq:tloss}
    \vspace{-1.5ex}
\end{equation}
where $\alpha$ increases with the number of training episodes, facilitating a gradual transition from imitation learning to reinforcement learning.

\vspace{-1ex}
\subsection{Cloud Server Design}
\vspace{-1ex}
In our approach, we use a cloud server to coordinate the training process among RL agents at different intersections. Specifically, for real traffic environments with extremely limited resources, the cloud server first sends pruned initial models that meet the requirements to each intersection. During the training process, the cloud server facilitates the knowledge sharing, by aggregating gradient information from these heterogeneous models, enabling effective collaboration among agents at different intersections.

\paragraph{Model Pruning.}
To meet resource requirements, we employ structured pruning at initialization. This technique leverages the concept that a randomly initialized dense network contains a subnetwork (referred to as a "winning ticket") capable of achieving performance comparable to the original dense network \cite{frankle2018lottery,kim2023parameter}. Specifically, for a dense network with parameters $\theta$, network pruning results in a new model $\theta \odot M$, where $M=\{0,1\}^{|\theta|}$ is a binary mask used for the pruning, and $\odot$ denotes the Hadamard product (element-wise multiplication). In our approach, we generate multiple Actor and Critic subnetworks for each intersection, applying a fixed set of pruning ratios to different network layers. As illustrated in the left part of Figure~\ref{fig:detail}, we create three pruned submodels from the base model for different intersections. In these submodels, lighter colors indicate pruned neurons, while darker colors represent retained neurons.

\paragraph{Knowledge Sharing.}
As a specialized form of supervised learning, imitation learning also requires a substantial number of samples to train RL agents effectively. To enhance learning efficiency and maximize the use of trajectory samples, we introduce a knowledge-sharing mechanism that aggregates gradients from heterogeneous submodels. Specifically, since the submodels for each intersection are derived from the same base model, we aggregate their gradients using a weighted average operation as follows:
\vspace{-1ex}
\begin{equation}
    \overline{\nabla L}=\frac{\sum^{N}_{i=1}\nabla L_i}{\sum^{N}_{i=1}M_i},
    \label{eq:sharing}
    \vspace{-0.5ex}
\end{equation}
where $N$ is the number of intersections, $\nabla L_i$ and $M_i$ represent the gradient of the loss function and the binary mask from the $i$-th intersection's submodel, respectively.

As shown in the left part of Figure~\ref{fig:detail}, each colored neural network represents different agents’ subnetworks, where the generated subnetworks share some subsets of parameters across multiple agents. In the aggregated model, each neuron is colored with the colors of agents who share the corresponding neuron.

\setlength{\floatsep}{0cm}
\begin{algorithm}[t]\scriptsize
\caption{\textbf{Training Procedure of FitLight}}
\label{alg1}
\textbf{Input}: i) episodes $S$;
ii) episode steps $T$;
iii) trajectory memory $M_T$;
iv) Actor model $\theta_A$;
v) Critic model $\theta_C$;
vi) expert $E$;
vii) batch size $B$.
\\
\textbf{Output}: i) $\theta_C$; ii) $\theta_A$.
\begin{algorithmic}[1] 

\STATE receive the network structures from the cloud server;
\STATE randomly initialize $\theta_A$ and $\theta_C$;
\FOR{$episode=1,2,\cdots,S$}
    \FOR{$step=1,2,\cdots,T$}
        \STATE obtain the current traffic state $s$ of the intersection;
        \STATE choose the action $a$ based on $s$;
        \STATE obtain the expert behavior $a_e$ from $E$;
        \STATE execute action $a$ at the intersection;
        \STATE observe the next state $s'$ of the intersection;
        \STATE store the trajectory $\langle s,a,a_e,r,s'\rangle$ in $M_T$;
        \IF{$Size(M_T)\geq B$}
            \STATE sample a mini-batch $b$ of size $B$ from $M_T$;
            \STATE compute $L_C$ by Equation~\ref{eq:closs};
            \STATE compute $L_A$ by Equation~\ref{eq:aloss};
            \STATE compute $L_I$ by Equation~\ref{eq:iloss};
            \STATE compute $L$ by Equation~\ref{eq:tloss};
            \STATE update model parameters $\theta_C$ and $\theta_A$;
            \STATE upload the gradient $\nabla L$ to the cloud server for aggregation by Equation~\ref{eq:sharing};
            \STATE receive $\overline{\nabla L}$ from the cloud server to update model parameters $\theta_C$ and $\theta_A$;
        \ENDIF
    \ENDFOR
\ENDFOR
\STATE {\bf return} trained models $\theta_C$ and $\theta_A$;

\end{algorithmic}
\end{algorithm}
\setlength{\textfloatsep}{0cm}

\begin{table*}[t]
\scriptsize
\centering
\caption{Comparison of average travel time.}
\vspace{-2ex}
\label{tab:average_travel_time}
\begin{tabular}{|c|c|ccccccccc|}
\hline
\multirow{3}{*}{\textbf{Type}}   & \multirow{3}{*}{\textbf{Method}} & \multicolumn{9}{c|}{\textbf{Average Travel Time (seconds)}}                                                                                                                               \\ \cline{3-11} 
                                 &                                  & \multicolumn{4}{c|}{\textbf{Synthetic Dataset}}                                           & \multicolumn{5}{c|}{\textbf{Real-world Dataset}}                                              \\ \cline{3-11} 
                                 &                                  & \textbf{Syn1}  & \textbf{Syn2}   & \textbf{Syn3}   & \multicolumn{1}{c|}{\textbf{Syn4}}   & \textbf{Hangzhou1} & \textbf{Hangzhou2} & \textbf{Jinan1} & \textbf{Jinan2} & \textbf{Jinan3} \\ \hline
\multirow{3}{*}{\textbf{Non-RL}} & \textbf{FixedTime}               & 380.35         & 453.73          & 534.47          & \multicolumn{1}{c|}{606.52}          & 525.28             & 537.82             & 444.84          & 378.41          & 403.22          \\
                                 & \textbf{MaxPressure}             & 122.68         & 162.32          & 245.26          & \multicolumn{1}{c|}{310.37}          & 404.67             & 456.11             & 373.76          & 371.24          & 356.30          \\
                                 & \textbf{MaxHP}                   & 122.98         & 159.72          & 211.72          & \multicolumn{1}{c|}{267.38}          & 362.34             & 419.82             & 330.00          & 330.04          & 320.77          \\ \hline
\multirow{12}{*}{\textbf{RL}}    & \textbf{PressLight}              & 108.23         & 145.43          & 186.28          & \multicolumn{1}{c|}{260.41}          & 351.55             & 425.61             & 305.21          & 302.56          & 294.08          \\
                                 & \textbf{A2C}                     & 117.60         & 133.65          & 236.58          & \multicolumn{1}{c|}{433.85}          & 339.24             & 416.08             & 375.12          & 322.92          & 288.92          \\
                                 & \textbf{FedLight}                & 108.71         & 136.45          & 172.45          & \multicolumn{1}{c|}{217.58}          & 341.94             & 410.40             & 290.38          & 290.58          & 278.69          \\
                                 & \textbf{PPO}                     & 105.86         & 138.11          & 204.76          & \multicolumn{1}{c|}{314.84}          & 358.66             & 421.89             & 321.37          & 304.66          & 286.43          \\
                                 & \textbf{InitLight}            & 102.86         & 126.44          & 172.36          & \multicolumn{1}{c|}{237.52}          & 333.80             & \textbf{374.60}    & 297.58          & 293.81          & 284.24          \\
                                 \cline{2-11}
                                 & \textbf{FairLight}               & 96.49          & 121.67          & 156.55          & \multicolumn{1}{c|}{198.52}          & 316.28             & \textbf{365.20}    & \textbf{262.55}          & \textbf{257.38}          & \textbf{250.69}          \\
                                 & \textbf{IPDALight}               & \textbf{89.66} & \textbf{110.00} & \textbf{146.34} & \multicolumn{1}{c|}{\textbf{182.72}} & \textbf{299.37}    & 403.18             & 264.31          & \textbf{256.89} & \textbf{253.41} \\
                                 \cline{2-11}
                                 & \textbf{FairLight(c)}           & 98.36          & 132.48          & 193.20          & \multicolumn{1}{c|}{228.59}          & 314.38             & 383.38             & 281.71          & 277.09          & 270.45          \\
                                 & \textbf{IPDALight(c)}           & 97.26          & 121.78          & 161.18          & \multicolumn{1}{c|}{205.31}          & 310.30             & 392.15             & 272.67          & 270.59          & 263.20          \\
                                 & \textbf{FitLight(p)}            & 96.13          & 123.62          & 160.23          & \multicolumn{1}{c|}{213.52}          & 339.15             & 441.40             & 317.18          & 301.83          & 287.60          \\
                                 & \textbf{FitLight(mp)}                & \textbf{90.41} & \textbf{112.98} & \textbf{149.05}          & \multicolumn{1}{c|}{\textbf{189.40}}          & \textbf{306.39}             & 378.84             & 261.82 & 261.50          & 253.83 \\ 
                                 \cline{2-11}
                                 & \textbf{FitLight}            & 91.06          & 113.57          & 150.83 & \multicolumn{1}{c|}{189.72} & 306.44    & 378.00             & \textbf{260.84} & 260.96 & 253.64          \\\hline
\end{tabular}
\vspace{-5ex}
\end{table*}

\subsection{FitLight Implementation}
Algorithm~\ref{alg1} details the training process of a FitLight agent. In lines 1-2, the agent is initialized with the pruned structure. Lines 5-10 show the interaction between the agent and the intersection, where the exact algorithm gives the label $a_e$ for the current state $s$. When the agent stores enough trajectory samples, lines 12-16 update the parameters of the PPO agent by using the local loss. Lines 18-19 show our knowledge-sharing mechanism, where the cloud server collects the gradient from all intersections and then dispatches the aggregated gradient to them for parameter update.

\section{Performance Evaluation}
\label{experiment}
To evaluate the effectiveness of our approach, we conduct experiments on a Ubuntu server equipped with an Intel Core i9-12900K CPU, 128GB memory, and NVIDIA RTX 3090 GPU. We implement our FitLight approach on an open-source traffic simulator, Cityflow \cite{zhang2019cityflow}, by using Python. During the simulation of traffic scenarios, similar to prior work \cite{wei2019presslight,ye2021fedlight,ye2023initlight}, we set the phase duration to 10 seconds. In FitLight, the base model of the PPO agent contains two neural networks, i.e., the Actor model with three layers (containing 13, 32, and 8 neurons) and the Critic model with three layers (containing 13, 32, and 1 neurons). We use the Adam optimizer for parameter updating and set the learning rate $\eta$ of Actor and Critic models to 0.0005 and 0.001, respectively. By default, we set the discount factor $\gamma$ to 0.99, the GAE parameter $\lambda$ is set to 0.95, the batch size $N$ of the data sampled for training to 5,  the clipping parameter $\epsilon$ to 0.2, and the balance factor $\alpha$ to $0.001\times \#$ $of$ $episode$.

We design comprehensive experiments to answer the following four research questions:

\textbf{RQ1 (Effectiveness):} Can FitLight explore a better TSC strategy starting from imitation learning? 
 
\textbf{RQ2 (Efficiency and Generalizability):} Can FitLight be plug-and-played in any traffic environment?

\textbf{RQ3 (Benefits):} Why can FitLight improve the learning performance and generalizability of RL models?

\textbf{RQ4 (Applicability):} Can FitLight be deployed in real-world embedded scenarios with extremely limited resources?

\paragraph{Baselines.}
Since we set the duration of the control phase to a constant value of $10s$, for a fair comparison, we chose eight representative baseline methods with constant duration, including three Non-RL methods and five RL-based methods as follows: i) \textbf{FixedTime} \cite{koonce2008traffic}; ii) \textbf{MaxPressure} \cite{varaiya2013max}; iii) \textbf{MaxHP}, our proposed expert algorithm; iv) \textbf{PressLight} \cite{wei2019presslight}; v) \textbf{A2C} \cite{ye2021fedlight}; vi) \textbf{FedLight} \cite{ye2021fedlight}; vii) \textbf{PPO} \cite{schulman2017proximal}, and; viii) \textbf{InitLight} \cite{ye2023initlight}. On the other hand, to further evaluate the performance of our approach, we also conduct two state-of-the-art dynamic duration-based methods for comparison: i) \textbf{FairLight} \cite{ye2022fairlight}, and; ii) \textbf{IPDALight} \cite{zhao2022ipdalight}. Moreover, to validate the effectiveness of our proposed approach, we compare with four ablation methods as follows: i) \textbf{FairLight(c)} \cite{ye2022fairlight}, a constant duration version of FairLight; ii) \textbf{IPDALight(c)} \cite{zhao2022ipdalight}, a constant duration version of IPDALight; iii) \textbf{FitLight(p)}, a pressure-based method that replaces the hybrid pressure in the sate representation and reward design of FitLight with pressure, and; iv) \textbf{FitLight(mp)}, the light version of the FitLight model pruning. For FitLight(mp), to meet the extremely resource-constrained requirements of \cite{xing2022tinylight} (i.e., a micro-controller with merely $16$ $KB$ RAM and $32$ $KB$ ROM), we set the pruning rate of each layer of the model to 0.2, 0.4, and 0.6, respectively. Under this setting, the memory cost of each PPO agent is only $14.83$ $KB$.

\paragraph{Datasets.}
We consider nine public multi-intersection datasets provided by \cite{wei2019survey}. For all datasets,  each intersection of all the road networks has four incoming roads and four outgoing roads, where each road has three lanes, i.e., turning left, going straight, and turning right. The details of the datasets are as follows:

\begin{itemize}
    \item \textbf{Synthetic datasets}: Four synthetic datasets, Syn1-4, contain $1\times 3$, $2\times 2$, $3\times 3$, and $4\times 4$ intersections, respectively. The vehicle arrival rates are modeled using a Gaussian distribution, with an average rate of 500 vehicles per hour for each entry lane.
    
\item \textbf{Real-world datasets}: 
Five real-world datasets (i.e., Hangzhou1, 2, and Jinan1-3) were collected using cameras deployed in the Gudang sub-district of Hangzhou and the Dongfeng sub-district of Jinan. Each dataset from Hangzhou contains 16 intersections arranged in a $4\times 4$ grid, while each dataset from Jinan includes 12 intersections arranged in a $3\times 4$ grid.
\end{itemize}

\begin{table*}[t]
\scriptsize
\tabcolsep=0.1cm
\centering
\caption{Comparison of convergence performance for different TSC methods.}
\vspace{-2ex}
\label{tab:first_episode_performance}
\begin{tabular}{|c|ccccccccc|}
\hline
\multirow{3}{*}{\textbf{Method}} & \multicolumn{9}{c|}{\textbf{Average Travel Time (seconds) / Start Episode \# of Converge (\#)}}                                                                                                                          \\ \cline{2-10} 
                                 & \multicolumn{4}{c|}{\textbf{Synthetic Dataset}}                                                            & \multicolumn{5}{c|}{\textbf{Real-world Dataset}}                                                            \\ \cline{2-10} 
                                 & \textbf{Syn1}       & \textbf{Syn2}       & \textbf{Syn3}       & \multicolumn{1}{c|}{\textbf{Syn4}}       & \textbf{Hangzhou1}  & \textbf{Hangzhou2}  & \textbf{Jinan1}     & \textbf{Jinan2}     & \textbf{Jinan3}     \\ \hline
\textbf{PressLight}              & 487.54 (79)         & 532.26 (123)        & 759.62 (142)        & \multicolumn{1}{c|}{781.93 (157)}        & 472.46 (80)         & 521.24 (61)         & 541.44 (75)         & 512.36 (76)         & 543.18 (82)         \\
\textbf{A2C}                     & 871.73 (95)         & 1306.89 (165)       & 1032.85 (N/A)       & \multicolumn{1}{c|}{1315.05 (N/A)}       & 1241.92 (122)       & 765.53 (65)         & 1298.90 (N/A)       & 1207.52 (105)       & 1224.99 (133)       \\
\textbf{FedLight}                & 916.81 (53)         & 1175.63 (111)       & 1309.26 (133)       & \multicolumn{1}{c|}{1357.66 (59)}        & 1009.92 (48)        & 807.72 (52)         & 1152.66 (47)        & 1213.16 (110)       & 1229.11 (70)        \\
\textbf{PPO}                     & 873.82 (83)         & 1056.83 (165)       & 1127.96 (145)       & \multicolumn{1}{c|}{1297.66 (N/A)}       & 813.43 (111)        & 694.34 (100)        & 972.88 (110)        & 1031.68 (89)        & 869.87 (80)         \\
\textbf{InitialLight}            & 115.73 \textbf{(1)}         & 164.63 \textbf{(2)} & 202.51 \textbf{(2)} & \multicolumn{1}{c|}{\textbf{262.45} (5)} & \textbf{330.78 (1)} & \textbf{387.06 (1)} & 300.42 \textbf{(1)}          & 294.18 \textbf{(1)}          & 288.42 \textbf{(1)}          \\
\hline
\textbf{FairLight}               & 530.37 (30)         & 686.65 (39)         & 876.40 (76)         & \multicolumn{1}{c|}{979.99 (58)}         & 517.50 (11)         & 512.99 (11)         & 703.27 (18)         & 677.06 (16)         & 588.71 (14)         \\
\textbf{IPDALight}               & 228.7 \textbf{(3)}           & 237.61 \textbf{(3)}          & 415.12 (14)         & \multicolumn{1}{c|}{420.39 \textbf{(3)}}          & 345.61 \textbf{(2)}          & 419.51 (23)         & 358.12 (10)         & 299.72 \textbf{(2)}          & 313.30 \textbf{(2)}          \\
\hline
\textbf{FairLight(c)}           & 609.42 (92)         & 774.53 (167)        & 1003.88 (N/A)       & \multicolumn{1}{c|}{1030.18 (N/A)}       & 721.58 (35)         & 588.67 (29)         & 757.77 (103)        & 691.84 (99)         & 687.28 (92)         \\
\textbf{IPDALight(c)}           & 592.45 (13)         & 819.53 (14)         & 979.09 (11)         & \multicolumn{1}{c|}{951.04 (11)}         & 558.56 (3)          & 533.64 (12)         & 711.45 (3)          & 634.07 (3)          & 672.85 (12)         \\
\textbf{FitLight(p)}            & 173.23 (4)          & 275.91 (6)          & 475.40 (7)          & \multicolumn{1}{c|}{630.18 (8)}          & 369.91 (8)          & 423.10 \textbf{(1)}          & 382.60 (6)          & 357.85 (6)          & 363.30 (7)          \\
\textbf{FitLight(mp)}                & \textbf{104.74 (1)}          & \textbf{136.94 (2)}          & \textbf{183.86 (2)}          & \multicolumn{1}{c|}{310.18 \textbf{(2)}}          & 330.88 \textbf{(2)}          & 392.23 \textbf{(1)}          & \textbf{281.88 (2)} & \textbf{282.30 (2)} & \textbf{274.37 (2)} \\
\hline
\textbf{FitLight}            & \textbf{99.21 (1)} & \textbf{125.05 (2)} & \textbf{163.77 (2)} & \multicolumn{1}{c|}{\textbf{215.21 (2)}} & \textbf{321.58 (1)} & \textbf{387.98 (1)} & \textbf{272.97 (2)} & \textbf{274.22 (1)} & \textbf{267.41 (1)} \\ \hline
\end{tabular}
\vspace{-5ex}
\end{table*}

\vspace{-1ex}
\subsection{Results of the Control Performance (RQ1)}
To answer RQ1, we compared FitLight against the fourteen baseline methods, in terms of average travel time, where we trained all RL-based methods using 200 episodes. Table~\ref{tab:average_travel_time} shows experimental results for different control methods. For each dataset, the TSC methods with the best or second-best performance are highlighted in bold. From this table, MaxHP can achieve a shorter average travel time than MaxPressure and competitive results compared with some RL-based methods, especially for larger datasets. These results show the effectiveness of our proposed hybrid pressure and illustrate the reason why we chose MaxHP as the expert algorithm. In this table, FitLight can outperform all constant duration methods. This is because, due to our proposed knowledge sharing mechanism, FitLight enables RL agents to jointly explore the optimal control strategy. On the other hand, due to the full use of duration, dynamic duration methods are significantly better than constant methods. However, our FitLight methods can still achieve a similar level with these two dynamic duration methods, where the biggest gap is only 3.51\%. Compared with the original uncompressed model, the perfomance decrease of FitLight(mp) is negligible, showing the practicality of our model pruning approach.

\begin{figure}[t]
    \centerline{\includegraphics[width=0.9\columnwidth]{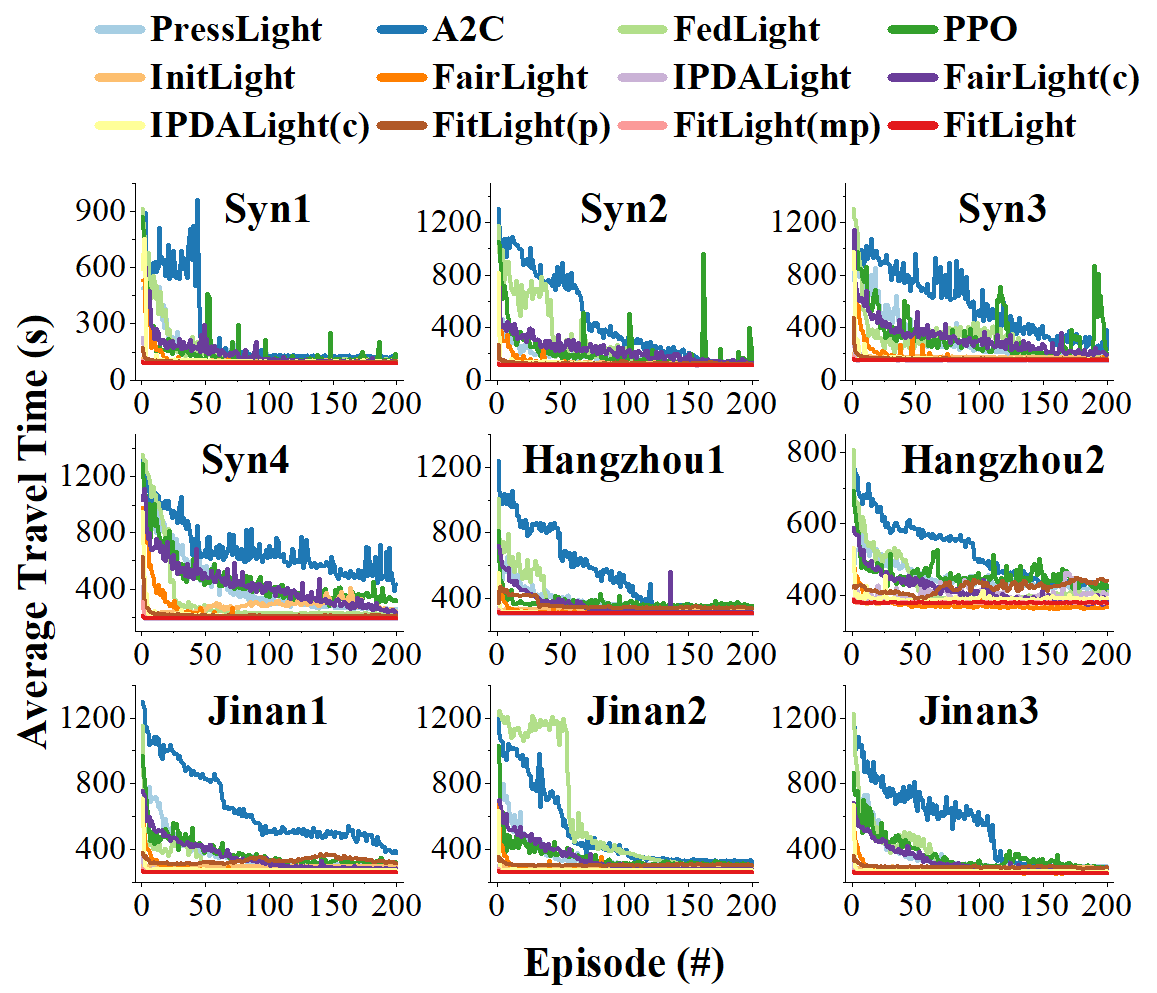}}
    \vspace{-2.5ex}
    \caption{Comparison of convergence rates.}
    \label{fig:convergence}
\end{figure}

\vspace{-1ex}
\subsection{Results of the Convergence (RQ2)}
\vspace{-1ex}
To evaluate the efficiency and the generalization ability of FitLight, Figure~\ref{fig:convergence} evaluates the convergence performance of the RL-based methods on the nine datasets. Compared to all baseline methods, our FitLight method achieves almost the lowest average travel time at the beginning of RL training for all the datasets with much fewer fluctuations. This is because our proposed knowledge sharing mechanism enables RL agents to perform effective imitation learning with only very few trajectory samples. In addition, although InitLight uses a pre-trained model that can also perform well at the beginning and converge fast, our FitLight can always achieve the lowest average travel time for all datasets eventually. This result shows that due to a better initial solution obtained by imitation learning, the subsequent reinforcement learning process of FitLight enables RL agents to explore a better final result, which confirms the seamless transition between imitation learning and reinforcement learning of our method. Note that, for all the evaluated datasets, FitLight can converge within 2 episodes, significantly faster than most baseline methods. The above facts evidently reveal the efficiency and the generalizability of our FitLight approach.

To further illustrate the advantages of FitLight, Table~\ref{tab:first_episode_performance} provides the detailed convergence information for different RL-based methods, focusing on jumpstart performance (i.e., average travel time during the first episode) and the episode at which the convergence begins, where the best and the second-best results are highlighted in bold. Here, the criterion for the convergence is based on the method described in \cite{zhao2022ipdalight}. From this table, we can find that FitLight achieves the best jumpstart performance on all datasets and the fastest convergence on most datasets. Due to the pre-trained initial model, InitLight can converge faster on some datasets (e.g., Hangzhou1 and Jinan1-3). However, the gap is only 1 episode, and FitLight can achieve better control performance without any pre-training. Note that due to the model pruning, although the jumpstart performance of the pruned FitLight(mp) model has slightly decreased, it can still converge to a final result that is similar to the unpruned version. These results demonstrate the plug-and-play capability of our FitLight approach for arbitrary traffic scenarios.

\vspace{-1ex}
\subsection{Quality of the Reward Function (RQ3)}
\begin{figure}[t]
    \centerline{\includegraphics[width=0.9\columnwidth]{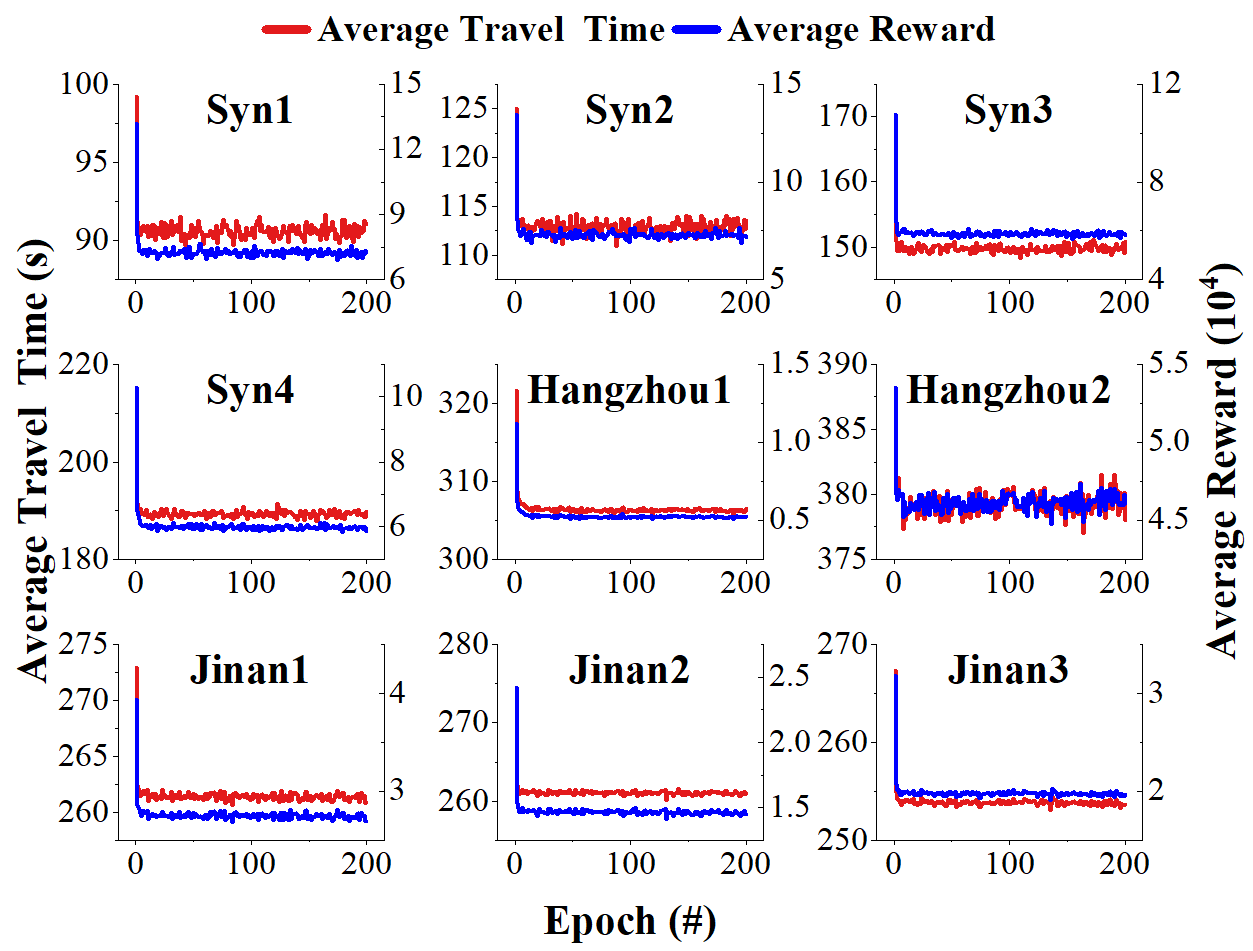}}
    \vspace{-2.5ex}
    \caption{Comparison of average travel time and reward.}
    \label{fig:reward}
\end{figure}

\vspace{-0.5ex}
In this paper, we use the newly proposed concept of hybrid pressure to represent the state and design the reward of the RL agent, which makes the agent take more traffic dynamics of individual vehicles into account. To justify our hybrid pressure design and understand why FitLight can plug and play for any datasets, as shown in Figure~\ref{fig:reward}, we compare the average travel time and the average reward of each episode on different datasets. From this figure, we observe that the average travel time is closely correlated with absolute values of the average reward (i.e., the average HP of intersections). These results support the effectiveness of our HP-based agent design in reducing the average travel time of vehicles. Moreover, the smooth change of rewards during training also confirms that our FitLight method supports a smooth transition from imitation learning to reinforcement learning. Thus, our FitLight can improve both the control performance and the generalization ability of RL agents.

\vspace{-1ex}
\subsection{Results of the Deployment Cost (RQ4)}
\vspace{-0.5ex}
To analyze the deployment cost of FitLight, we built a cloud-edge simulation platform consisting of the server mentioned above and 16 Raspberry Pi 4B boards (with ARM Cortex-A72 CPU and 2G RAM), where we deploy a FitLight agent on each Raspberry Pi board to simulate real-world intersection scenario. For memory cost, as mentioned above, due to the model pruning, the model size of each agent is only 14.83 $KB$. which can be deployed on most resource-constrained embedded systems. For computation cost, the agent only needs 0.05 $ms$ to make a control decision of phase selection. Within an autonomous TSC system, the agent also needs to update its model parameters after collecting some trajectory data. In our approach, an agent takes 5 samples from its trajectory memory at a time for model update, which costs around 51.67 $ms$. Compared with the 10-second signal phase duration, this inference and training costs can meet the real-time requirements of embedded devices in most real-world scenarios. On the other hand, since FitLight includes a federated learning-based knowledge-sharing mechanism, we also evaluate the communication cost of our approach. In the experiment, we set the phase duration of traffic lights to 10 seconds, which means that every 10 seconds, the agent of an intersection needs to interact with its traffic environment and store a trajectory sample. Once the agent collects a batch of 5 new samples, it will use these samples to calculate and share the gradient. In other words, every 50 seconds, each edge device needs to send and receive 14.83 $KB$ gradients, respectively. This communication cost is tolerable for most IoT devices, since in each hour, one device only has a communication overhead of 2.09 $MB$.

\section{Conclusion}
\label{conclusion}
Due to trial-and-error attempts during the training process, existing RL-based TSC methods suffer from the problem of high learning costs and poor generalizability. To address this problem, in this paper, we propose a novel FIL-based approach named FitLight, which enables RL agents to be plug-and-play for any traffic scenarios. Based on the proposed federated imitation learning frameworks and hybrid pressure-based agent design, our FitLight agent can smoothly transition from imitation learning to reinforcement learning. Therefore, the RL agent can quickly find a high-quality initial solution and then find a better final control strategy. Experimental results on various well-known benchmarks show that, compared with the state-of-the-art RL-based TSC methods, FitLight can not only converge faster to competitive results, but also exhibit stronger robustness in different traffic scenarios. Especially, our approach can achieve near-optimal performance in the first episode.

\clearpage
\balance
\bibliographystyle{IEEEtran}
\bibliography{aaai25}

\end{document}